\documentclass[runningheads]{llncs}
\usepackage[utf8]{inputenc} 
\usepackage[T1]{fontenc}    
\usepackage{url}            
\usepackage{booktabs}       
\usepackage{amsfonts}       
\usepackage{nicefrac}       
\usepackage{microtype}      

\usepackage{algorithmicx}
\usepackage{algorithm} 
\usepackage[noend]{algpseudocode} 
\usepackage{amsmath}
\usepackage{amssymb}
\usepackage[pdftex]{graphicx}
\usepackage{multirow}
\usepackage{footnote}
\usepackage{wrapfig}
\usepackage[colorlinks=true,urlcolor=black,linkcolor=blue,citecolor=blue]{hyperref}

\usepackage[capitalize,nameinlink]{cleveref}
\usepackage{tablefootnote}
\usepackage{xcolor}
\usepackage{enumitem}
\usepackage{relsize}
\usepackage{subcaption}
\usepackage{float}
\usepackage{mathrsfs}
\usepackage{longtable}

\makeatletter
\AtBeginDocument{%
  \def\doi#1{\url{https://doi.org/#1}}}
\makeatother

\begin{document}
\title{PEREGRiNN: Penalized-Relaxation Greedy Neural Network Verifier}
%
\author{Haitham Khedr \and James Ferlez\and Yasser Shoukry}
\authorrunning{Khedr et al.}

%
\institute{
University of California, Irvine, USA
}
\maketitle              
\begin{abstract}
Neural Networks (NNs) have increasingly apparent safety implications 
commensurate with their proliferation in real-world applications: both 
unanticipated as well as adversarial misclassifications can result in fatal 
outcomes. As a consequence, techniques of formal verification have been 
recognized as crucial to the design and deployment of safe NNs. In this paper, 
we introduce a new approach to formally verify the most commonly considered 
safety specifications for ReLU NNs -- i.e. polytopic specifications on the 
input and output of the network. Like some other approaches, ours uses a 
relaxed convex program to mitigate the combinatorial complexity of the problem.
However, unique in our approach is the way we use a convex solver not only as a 
linear feasibility checker, but also as a means of penalizing the amount of 
relaxation allowed in solutions. In particular, we encode each ReLU by means of 
the usual linear constraints, and combine this with a convex objective function 
that penalizes the discrepancy between the output of each neuron and its 
relaxation. This convex function is further structured to force the largest 
relaxations to appear closest to the input layer; this provides the further 
benefit that the most ``problematic'' neurons are conditioned as early as 
possible, when conditioning layer by layer.
This paradigm can be leveraged to create a verification algorithm that is not 
only faster in general than competing approaches, but is also able to verify 
considerably more safety properties; we evaluated PEREGRiNN on a standard MNIST 
robustness verification suite to substantiate these claims.
%

\end{abstract}
%
%
%

\section{Introduction}
\label{sec:intro}

Neural Networks have become an increasingly central component of modern machine 
learning systems, including those that are used in safety-critical 
cyber-physical systems such as autonomous vehicles. The rate of this adoption 
has exceeded the ability to reliably verify the safe and correct functioning of 
these components, especially when they are integrated with other components 
such as controllers. Thus, there is an increasing need to verify that NNs 
reliably produce safe outputs, especially subject to malicious adversarial
inputs~\cite{szegedy2013intriguing,goodfellow2014explaining,kurakin2016adversarial,song2018physical}.

In this paper, we propose PEREGRiNN, an algorithm for efficiently and formally 
verifying the input/output behavior of ReLU NNs.
In this context, PEREGRiNN falls into the broad category of sound and complete 
\emph{search and optimization} NN verifiers 
\cite{LiuAlgorithmsVerifyingDeep2019}. The \emph{search} aspect of PEREGRiNN 
involves iterating over different combinations of neuron activation patterns to 
verify that each is compatible with the specified safety constraints (on the 
input and output of the network). Like other algorithms in this category, 
PEREGRiNN combines this search with \emph{optimization} techniques to make 
inferences about the feasibility of full-network activation patterns on the 
basis of activation patterns of only a subset of neurons. The optimization in 
question reformulates the original NN feasibility problem into a relaxed convex 
feasibility problem to allow sound inferences: i.e. if the convex relaxation is 
infeasible, then the original NN problem may soundly be concluded to be 
infeasible. In this relaxed feasibility problem, the output of each individual 
neuron is assigned a relaxation variable that is decoupled from the actual 
output of that neuron. 
PEREGRiNN also 
uses a type of reachability analysis (symbolic interval analysis) both to 
enhance the optimization-based inference described above and as a source of 
additional sound inference itself. For this reason, PEREGRiNN's search 
procedure searches neurons in a layer-by-layer fashion, preferring to fix the 
phases of neurons closest to the input layer first.

In contrast to other search and optimization algorithms, however, PEREGRiNN 
\emph{augments} each convex feasibility query with a (convex) penalty function 
in order to obtain better guidance on which activation patterns to search next. 
In particular, we note that the amount of relaxation needed on a neuron can be regarded as 
a \emph{quasi-measure} of how close the convex solver came to operating the 
associated neuron in a valid regime -- i.e. at a valid evaluation of that 
neuron on a particular input. In this sense, the amount of relaxation in 
aggregate can be regarded as a quasi-measure of how close the solver came to 
finding a valid evaluation of the network as a whole. Inversely, the largest 
distance between a relaxation variable and its neuron's closest ReLU constraint 
intuitively corresponds in some sense to how ``problematic'' that neuron is 
with regard to obtaining such a valid evaluation. These distances we refer to 
as the \emph{``slacks''} for each neuron. Thus, PEREGRiNN may be 
regarded as \emph{greedily} minimizing a \emph{slack-based penalty}.

Finally, we evaluated the performance of PEREGRiNN by using it to verify the 
adversarial robustness of networks trained on the MNIST~\cite{mnist} dataset. 
Our experiments show that PEREGRiNN is on average 1.4$\times$ faster than 
Neurify~\cite{wang2018efficient}, 1.35$\times$ faster than 
Venus~\cite{botoeva2020efficient}, 1.25$\times$ faster than 
nnenum~\cite{bak2020improved}, and 1.85$\times$ faster than 
Marabou~\cite{katz2019marabou}. It also proves 40 \%, 30 \%, 20\%, and 65 \%  
more properties than the other solvers, respectively. PEREGRiNN's unique convex 
penalty augmentations are also considered in ablation experiments to validate 
their  benefits.

\paragraph{Related work.} %
Since PEREGRiNN is a sound and complete verification algorithm, we restrict our 
comparison to other sound and complete algorithms. NN verifiers can be grouped 
into roughly three categories: (i) SMT-based methods, which encode the problem 
into a Satisfiability Modulo Theory 
problem~\cite{katz2019marabou,KatzReluplexEfficientSMT2017a,ehlers2017formal}; 
(ii) MILP-based solvers, which directly encode the verification problem as a 
Mixed Integer Linear 
Program~\cite{lomuscio2017approach,tjeng2017evaluating,bastani2016measuring,bunel2020branch,fischetti2018deep,anderson2020strong,cheng2017maximum,botoeva2020efficient}
;
(iii) Reachability based methods, which perform layer-by-layer reachability 
analysis to compute the reachable
set~\cite{bak2020improved,xiang2017reachable,xiang2018output,gehr2018ai2,wang2018formal,tran2020nnv,ivanov2019verisig,fazlyab2019efficient,BakImprovedGeometricPath2020};
and (iv) convex relaxations 
methods~\cite{wang2018efficient,dvijotham2018dual,wong2017provable}.
In general, (i), (ii) and (iii) suffer from poor scalability. On the other 
hand, convex relaxation methods depend heavily on pruning the search space of 
indeterminate neuron activations; thus, they generally depend on obtaining good 
approximate bounds for each of the neurons in order to reduce the search space 
(the exact bounds are computationally intensive to compute 
\cite{dutta2017output}). These methods are most similar to PEREGRiNN: for 
example, 
\cite{wang2018formal,bunel2020branch,royo2019fast} 
recursively refine the problem using input splitting, and 
\cite{wang2018efficient} does so via neuron splitting. Other search and 
optimization methods include: Planet \cite{ehlers2017formal}, which combines a 
relaxed convex optimization problem with a SAT solver to search over neurons' 
phases; and Marabou~\cite{katz2019marabou}, which uses a modified simplex 
algorithm. 


\section{Problem formulation}
\label{sec:problem_formulation}
In this paper, we will consider Rectified Linear Unit (ReLU) NNs. An $n$-layer 
ReLU network, is a composition of $n$ ReLU layer functions: i.e. $\mathcal{NN} 
= f_n \circ f_{n-1} \circ \dots \circ f_1$ where the $i^\text{th}$ ReLU layer 
function 
is defined as $f_{i} : y \in \mathbb{R}^{k_{i-1}} \mapsto \max\{ W_i y + b_i, 0 
\} \in \mathbb{R}^{k_i}$. We refer to $f_1$ as the \underline{input layer}.
Finally, to refer to individual neurons, we use the notation $(z)_j$ to 
indicate the $j^\text{th}$ element of $z$.

\textbf{Verification Problem.} Let $\mathcal{NN}$ be an $n$-layer NN as defined 
above. Furthermore, let $P_{y_0} \subset \mathbb{R}^{k_0}$ be a convex polytope 
in the input space of $\mathcal{NN}$, and let $P_{y_n} \subset 
\mathbb{R}^{k_n}$ be a convex polytope in the output space of $\mathcal{NN}$. 
Finally, let $h_\ell : \mathbb{R}^{k_0} \times \mathbb{R}^{k_n} \rightarrow 
\mathbb{R}$, $\ell = 1, \dots, m$ be convex functions defining joint 
input/output constraints on $\mathcal{NN}$.
Then the verification problem is to decide whether
\begin{equation}
\label{eq:main_problem}
	\left\{ x \in \mathbb{R}^{k_0} \; \Big\lvert \; x \in P_{y_0} \; \mathlarger{\mathlarger{\wedge}} \; \mathcal{NN}(x) \in P_{y_n} \; \mathlarger{\mathlarger{\wedge}} \; \Big(\overset{m}{\underset{\ell=1}{\wedge}} h_\ell(x, \mathcal{NN}(x)) \leq 0 \Big)  \right\} = \emptyset.
\end{equation}

\section{PEREGRiNN Overview} 
\label{sec:peregrinn_overview}
\begin{figure}[t] 
    \centering \includegraphics[width=0.85\textwidth]{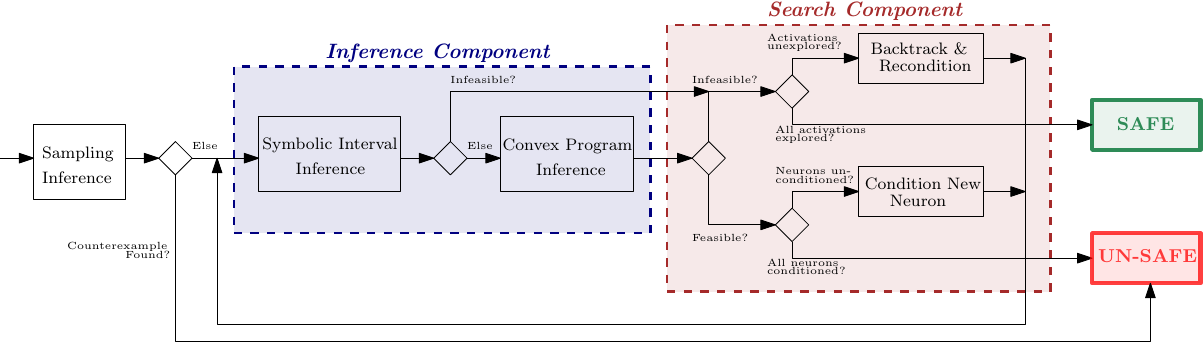}
\caption{Block Diagram of the PEREGRiNN Algorithm}
	\label{fig:peregrinn_block}
\end{figure}

The general structure of PEREGRiNN is depicted in \cref{fig:peregrinn_block}. 
Like other search and optimization based NN verifiers it has two main 
components: a \emph{search component} and an \emph{inference component}, and 
PEREGRiNN iterates back and forth between these these two components until 
termination. In particular, the search and inference components interact in the 
following way. The search component successively iterates over all possible 
on/off activations for each neuron; this is done by fixing these activations 
one neuron at a time, starting from the input layer and working towards the 
output layer. The process of fixing a neuron's activation is referred to as 
\emph{conditioning its phase}: each neuron can be in either its active phase 
(operating linearly) or inactive phase (outputting zero). Thus, the search 
component provides the inference component a subset of neurons, each of which 
has been conditioned; the inference component then attempts to soundly reason 
about whether the remaining, unconditioned neurons can be operated in such a 
way as to violate the safety constraint. If the inference component soundly 
concludes safety for all possible activations of the remaining unconditioned 
neurons, then the search component backtracks, oppositely reconditioning one of 
the neurons that was already conditioned. Otherwise, if a sound safe conclusion 
is not made, then the search component uses information from the inference 
component to decide on a new neuron to condition, and the process repeats. The 
algorithm terminates if either a counterexample to safety is found, or else all 
possible neuron activations are considered without finding such a 
counterexample. 

The convex program inference block is at the heart of the inference component 
and PEREGRiNN itself. In this block, PEREGRiNN, like other search and 
optimization solvers, uses a relaxed linear feasibility program where the 
output of each individual neuron is assigned a relaxation variable that is 
decoupled from the actual output of that neuron.
In the notation of \cref{sec:problem_formulation}, 
such a linear feasibility program can be written as follows, where the vector 
variables $y_i, i\neq0$ are the relaxation variables.
\begin{equation}\label{eq:generic_convex_program}
	\begin{cases}
		y_{i}\geq 0, \;
		y_i \geq W_i y_{i-1} + b_i &\forall i = 1, \dots, n \\
		y_0\in P_{y_0},\;
		y_n\in P_{y_n}^\mathsf{c}, \;
		\underset{\ell=1}{\overset{m}{\wedge}} h_\ell(y_0,y_n) \leq 0 & ~
	\end{cases}
\end{equation}
Importantly, if \eqref{eq:generic_convex_program} is infeasible, then the 
original NN problem in \eqref{eq:main_problem} may be soundly concluded to be 
infeasible as well -- and hence, safe. However, as described above, the primary 
function of the convex feasibility program is to use a set of conditioned 
neurons supplied by the search component in order to soundly reason about the 
remaining neurons. To do this, the conditioned neurons supplied by the search 
component are incorporated into the feasibility program 
\eqref{eq:generic_convex_program} as \emph{equality} constraints in the 
following way:
\begin{align}
	\text{Neuron } (y_i)_j \text{~ ON: }& \;\;  (y_{i})_j = (W_i y_{i-1} + b_i)_j \wedge (y_{i})_j \geq 0 \label{eq:convex_constraint_1}\\
	\text{Neuron } (y_i)_j \text{ OFF: }& \;\;  (y_{i})_j = 0 \wedge (W_i y_{i-1} + b_i)_j \leq 0.
	 \label{eq:convex_constraint_2}
\end{align}
Inferences created by the symbolic interval inference block using Symbolic 
Interval Analysis \cite{wang2018formal} are also incorporated using equality 
constraints like \eqref{eq:convex_constraint_1} and 
\eqref{eq:convex_constraint_2}.

Of the remaining blocks, the ``Backtracking \& Reconditioning'' block is 
essentially described above. The ``Condition New Neuron'' and ``Sampling 
Inference'' blocks have features unique to PEREGRiNN that are described in 
\cref{sec:peregrinn_enhancements}; the former implements a novel neuron 
prioritization, and the latter is a unique approach to quickly obtaining 
initial safety counterexamples.



\section{PEREGRiNN Enhancements} 
\label{sec:peregrinn_enhancements}
\subsection{Sum-of-Slacks Penalty} 
\label{sub:sum_of_slacks_penalty}
The core enhancement in PEREGRiNN is the inclusion of a specific objective 
function in the convex program used by the inference component. As per the 
discussion above, this objective function is interpreted as a \emph{penalty} on 
how far away a particular solution is from a valid input/output response of the 
network (and activation pattern on all hidden neurons). Specifically, this 
penalty function penalizes the sum of all of the ``slack'' variable for the 
entire network, where each neuron's slack variable is defined as $s_i 
\triangleq y_i - ( W_{i} \cdot y_{i-1} + b_{i})$. That is the distance between 
a relaxation variable $y_i$ and the linear response of its associated neuron.
During each feasibility/inference call, this has the obvious effect of 
incentivizing the convex solver to choose an actual input/output response of 
the network.

In addition, this penalty is effectively the $L_1$-norm of the \emph{vector} of all the slack variables, since the slack variables are non-negative. The $L_1$-norm of a vector, used as a penalty function, is well known to effectively encourage \emph{sparsity} on the resulting optimal solution. Thus, the sum-of-slacks effectively incentivizes the convex solver to leave as \emph{few} neurons as possible 
indeterminate in the solution. That is a sum-of-slacks penalty effectively encourages the convex solver to fix the phases of as many neurons as possible.


\subsection{Max-Slack Conditioning Priority} 
\label{sub:max_slack_conditioning_priority}
As noted above, the search component of PEREGRiNN operates layer-wise from 
input layer to output layer in order to leverage Symbolic Interval Analysis for 
additional inference. Hence, the search component always chooses the next 
neuron to be searched (i.e. conditioned) from among those as-yet-unconditioned 
neurons that are closest to the input layer. It further makes sense to only 
consider conditioning neurons that the convex solver was unable to operate at 
valid inputs/output. However, the convex solver typically returns several 
neurons to choose from with this property, and it is necessary to choose which 
of them to search next. Given the interpretation of a neuron's ``slack'' 
variable as a measure of how ``problematic'' that neuron was for the solver to 
obtain a valid evaluation of the network, PEREGRiNN's search component chooses 
the next neuron to condition based on slack-order ranking of those neurons that 
are not being operated at valid input/output points. This ``max-slack'' 
heuristic choice is unique to PEREGRiNN; compare to the output gradient 
heuristic employed in \cite{wang2018efficient}.

\subsection{Layer-wise-Weighted Penalty} 
\label{sub:layer_wise_weighted_penalty}
PEREGRiNN takes the ``max-slack'' neuron search priority one step further, 
though. Using techniques similar to those in \cite{ShoukrySMCSatisfiabilityModulo2018}, it is possible to show that there exists weights $q_1, \dots, q_n$ such that solving \eqref{eq:generic_convex_program} with the penalty 
\begin{equation}\label{eq:weighted_penalty}
    \min_{y_0,..,y_n} \sum_{i=0}^n\sum_{j=1}^{k_i}  q_i s_{ij}
\end{equation}
will result in a solution that is guaranteed to concentrate the most total slack in the earliest (unconditioned) layer. Thus, by using the layer-wise weighted sum-of-slacks penalty in \eqref{eq:weighted_penalty}, PEREGRiNN is uniquely able to force the (unconditioned) layer closest to 
the input layer to have the \emph{largest} total slack among all the layers. 
As a consequence, PEREGRiNN effectively concentrates the most ``problematic'' neurons in the 
layer where the next conditioning choice will be made. This 
scheme makes it much more likely that the neuron with the highest slack among 
\emph{all} of the neurons will be among the next neurons considered for 
conditioning -- in effect, often guiding the search component to condition on 
the most problematic neuron in the whole network (although this is not 
guaranteed).

\subsection{Initial Counterexample Search by Sampling} 
\label{sub:initial_counterexample_search_by_sampling}
Finally, PEREGRiNN incorporates a simple, yet novel, approach to help identify 
un-safe networks early on. In particular, PEREGRiNN uses the Volesti 
\cite{EmirisPracticalPolytopeVolume2018} Python library to uniformly sample 
points within the input constraint set, $P_x$, and evaluates the network 
exactly for those inputs. Unsafe outputs thus constitute counterexamples to the 
safety of the network, and PEREGRiNN can terminate.



\section{Experiments}
\label{sec:experiments}

We evaluated the performance and effectiveness of PEREGRiNN at verifying the 
adversarial robustness of NNs trained to recognize digits using the standard 
MNIST dataset. This verification problem fits into the general NN verification 
problem described in \cref{sec:problem_formulation}, and it is described 
subsequently in detail. In this context, we evaluated PEREGRiNN with two 
objectives described as follows.
\begin{enumerate}
	\item We conducted ablation experiments for all of PEREGRiNN's novel 
		features as described in \cref{sec:peregrinn_enhancements}. In 
		particular, we compared the performance of a full implementation of 
		PEREGRiNN -- i.e. \emph{exactly} as described in 
		\cref{sec:peregrinn_enhancements} -- with implementations that are 
		otherwise the same except for changing one and only one of the 
		following: the penalty function used in the convex program inference 
		block; the neuron prioritization used by the search component.

	\item We compared PEREGRiNN against other state-of-the-art NN verifiers, 
		both in terms of the time required to verify individual networks and 
		properties and in terms of the number of properties proved with a 
		common, fixed timeout.
\end{enumerate}

\noindent\textbf{Implementation.} We implemented PEREGRiNN in Python, and used 
an off-the-shelf Gurobi 9.1~\cite{optimization2013gurobi} convex optimizer for 
solving linear programs; the Volesti \cite{EmirisPracticalPolytopeVolume2018} Python 
interface was used to sample from the input polytope for the 
sampling inference block. For the other NN verifiers, 
we used publicly available implementations that were published by their 
creators (citations are included below). Each instance of 
of any verifier was run within its own single-core Virtual Box VM with 30 GB of 
memory. The VMs were run no more than 4 at a time on a host machine with 48 
hyperthreaded cores and 256 GB of memory; this ensured that each VM had 
adequate access to hardware resources.


\subsection{Adversarial Robustness Verification Task}
Subsequent experiments used the testbench we describe in this section.
\subsubsection{Neural Networks.} 
\label{ssub:neural_networks}
We used three ReLU NNs to recognize digits using the standard MNIST training 
database. The size and architectures of these networks are described in 
\cref{Models_Arch}. Each entry in the ``Architecture'' column of 
\cref{Models_Arch} describes the architecture of the corresponding network in 
terms of number of neurons per layer, from input layer on the left to output 
layer on the right.
%
\begin{table}[t]
\caption{Architecture of the NN models used in the experiments}
\label{Models_Arch}
\centering
\begin{tabular}{|c|c|c|c|c}
\hline
Models & \# ReLUs & Architecture\\
\hline
MNIST\_FC1 & 512 & \textless784,256,256,10\textgreater \\
\hline
MNIST\_FC2 & 1024 & \textless784,256,256,256,256,10\textgreater \\
\hline
MNIST\_FC3 & 1536 & \textless784,256,256,256,256,256,256,10\textgreater\\
\hline
\end{tabular}
\end{table}
%
\subsubsection{Verification Properties.} 
\label{ssub:verification_property}
We created a number of NN verification tasks based on proving whether the above 
described networks were robust against max-norm perturbations of their inputs. 
In particular, each verification task involves proving whether a particular 
input image, $x^\prime$, always results in the same classification when it is 
subjected to a $\max$-norm perturbation of at most some fixed size, $\epsilon > 
0$. Thus, each such verification problem is parameterized by both the specified 
input image, $x^\prime$, and the maximum amount of perturbation, $\epsilon$.

Formally, let $x^\prime$ be a given image in category $t \in \{1, \dots, M\}$, 
and let $\epsilon > 0$ be a specified maximum amount of $\max$-norm 
perturbation of $x^\prime$. Then we say that a NN with $M$ classification 
outputs, $\mathcal{NN}$, is robust
if for each classification category $m \in \{1, \ldots, M\}\setminus \{t\}$ the 
set of inputs yielding classification of $x^\prime$ as $m$
\begin{equation}\label{eq:adversarial_robustness_property}
\phi_m \triangleq \{x \mid x\in \mathbb{R}^{k_0}\text{, }\|x-x^{\prime}\|_\infty \leq \epsilon \text{, } z \in \mathbb{R}^{k_n}\text{, } \underset{i=1,\dots,n}{\max} \mathcal{NN}(x)_i = \mathcal{NN}(x)_m \} 
\end{equation}
is empty. Note that each instance of \eqref{eq:adversarial_robustness_property} 
is compatible with the problem in \eqref{eq:main_problem}.



\subsubsection{Adversarial Robustness Verifier Testbench} 
\label{ssub:adversarial_robustness_verifier_testbench}
Our verification testbench was then constructed by selecting 50 test images 
from the MNIST test dataset. Each test instance was then a combination of one 
of those images, one of the networks from \cref{Models_Arch} and one the 
following two $\max$-norm perturbations, $\epsilon = 0.02$ or $\epsilon = 
0.05$. 
Thus, each verification test in our testbench can be identified by one of 300 
tuples of the form:
	$(\text{\itshape net}, \text{\itshape image}, \text{\itshape perturb.}) \in \mathscr{T}\negthinspace\mathscr{B} \triangleq \{\text{\texttt{FC1}}, \text{\texttt{FC2}}, \text{\texttt{FC2}}\} \times \{1, \dots, 50\} \times \{0.02, 0.05\}$.

\subsection{Ablation Experiments} 
\label{sub:ablation_experiments}
In this series of experiments we evaluated the contribution that each of the 
primary PEREGRiNN enhancements made to its overall performance. This was done by 
comparing the full PEREGRiNN algorithm -- as described in 
\cref{sec:peregrinn_enhancements} -- with altered versions that replace exactly 
one of those enhancements at a time.

\begin{figure}[t]
    \centering
    \begin{subfigure}[t]{0.47\textwidth}
         \includegraphics[scale = 0.45,trim=1mm 0 0 0]{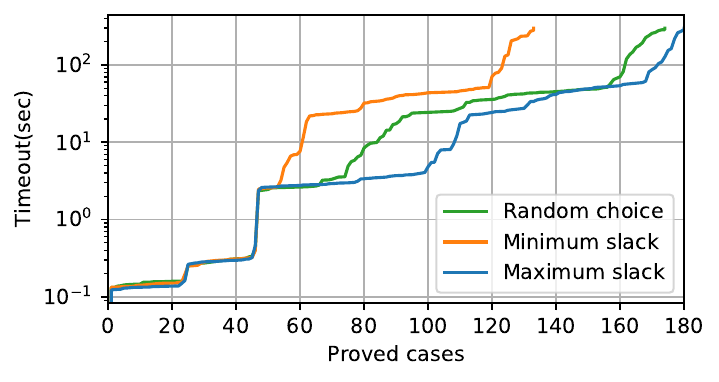}
    \caption{Cactus plot; proved cases vs. timeout}
    \label{fig:cactus_heuristics}
    \end{subfigure}
    \hfill
    \begin{subfigure}[t]{0.47\textwidth}
         \includegraphics[scale = 0.45,trim=2mm 0 0 0]{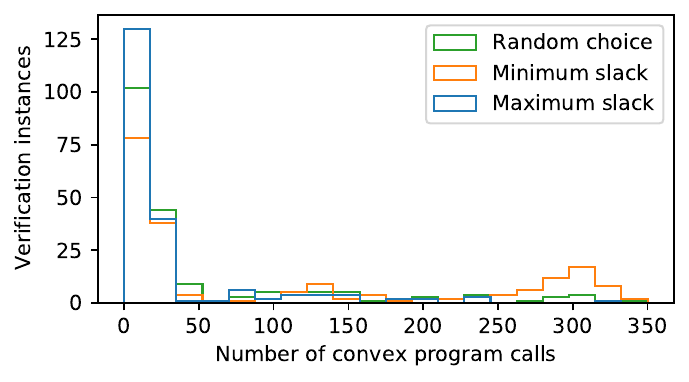}
    \caption{Histogram; number convex calls used}
    \label{fig:hist_heuristics}
    \end{subfigure}
    \caption{Performance of PEREGRiNN variants with different conditioning priorities}
    \label{fig:heuristics}
\end{figure}
\noindent\textbf{\itshape Note:} removing core features of PEREGRiNN often 
resulted in much longer run times, so the experiments in this section use a 
testbench $\mathscr{T}\negthinspace\mathscr{B}^\prime \subset 
\mathscr{T}\negthinspace\mathscr{B}$
that excludes all tests with one of the larger networks \texttt{FC2} or 
\texttt{FC3} \emph{and} $\epsilon = 0.05$.
\subsubsection{Penalty Function Ablation.} 
\label{ssub:penalty_function_ablation}
\begin{figure}[t]
    \centering
    \begin{subfigure}[t]{0.47\textwidth}
         \includegraphics[scale = 0.45,trim=0.7in 1in 1.3in 1.6in]{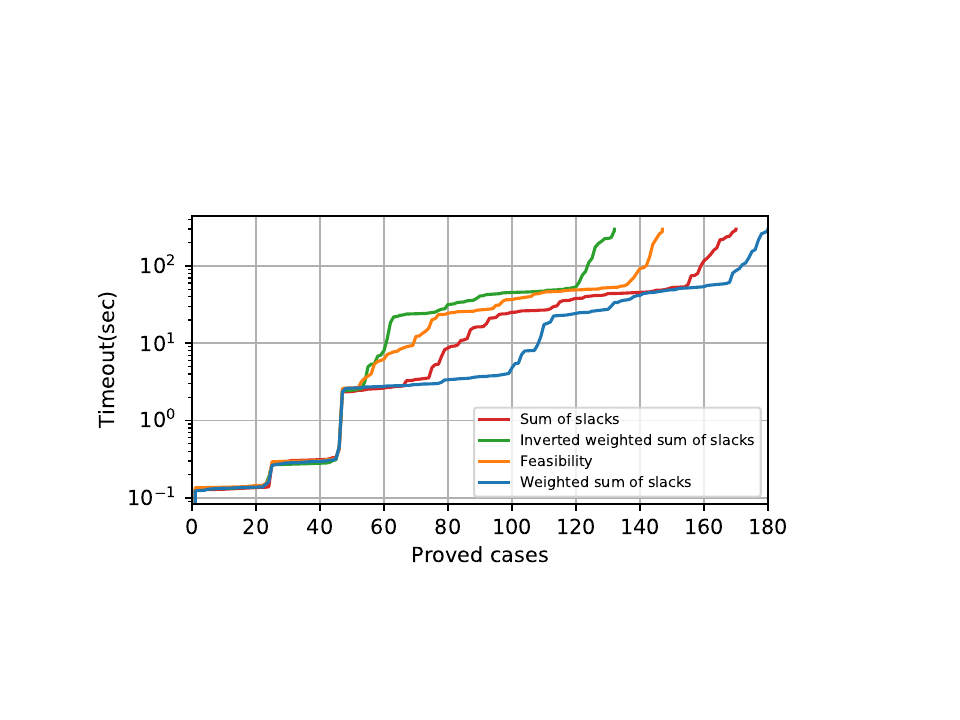}
    \caption{Cactus plot; proved cases vs. timeout}
    \label{fig:cactus_obj}
    \end{subfigure}
    \hfill
    \begin{subfigure}[t]{0.47\textwidth}
         \includegraphics[scale = 0.45,trim=1mm 0 0 0]{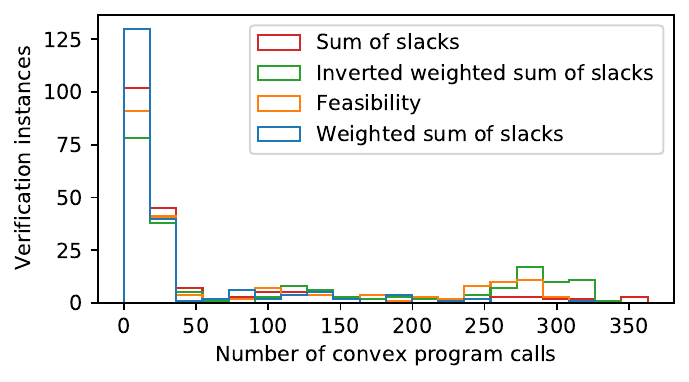}
    \caption{Histogram; number convex calls used}
    \label{fig:histogram_obj}
    \end{subfigure}
    \caption{Performance of PEREGRiNN variants with different objective functions}
    \label{fig:Objectives}
    \vspace{1mm}
\end{figure}

Our first ablation experiment evaluated the contribution of PEREGRiNN's unique 
penalty function features; see \cref{sub:sum_of_slacks_penalty} and 
\cref{sub:layer_wise_weighted_penalty}. In particular, we ran different 
variants of PEREGRiNN with the following penalty functions used inside the 
convex program inference block:
\vspace{-1.25mm}%
\begin{enumerate}
	\item {\itshape ``Weighted sum of slacks''}: PEREGRiNN's own weighted 
		sum of slacks penalty;

	\item {\itshape ``Sum of slacks''}: A sum-of-slacks penalty with equal 
		weighting on all layers;

	\item {\itshape ``Feasibility''}: A feasibility-only convex program such 
		as the one used in other tools, e.g. \cite{wang2018efficient} (i.e. 
		simply using a constant penalty function of 1);

	\item {\itshape ``Inverted weighted sum of slacks''}: PEREGRiNN's own 
		weighted sum of slacks penalty, except with the layer-wise weights 
		applied in reverse order force slack towards deeper layers rather than  
		shallower ones (see also \cref{sub:layer_wise_weighted_penalty}).
\end{enumerate}
\vspace{-1.25mm}%
\cref{fig:cactus_obj} shows a cactus plot of the number of proved cases vs. the 
timeout permitted to the algorithm: i.e. to prove at least a specified number 
of the test cases, each algorithm must have its timeout set at to the value of 
its curve in \cref{fig:cactus_obj}.
\cref{fig:histogram_obj} shows a histogram of the number of times each of the 
algorithm variants needed to call the convex solver in order to terminate; this 
quantifies each algorithm's cost in a well-known unit of computation, also the 
single most computationally costly part of PEREGRiNN. \cref{fig:histogram_obj} 
plots the number of convex solver calls required for evenly spaced bins of 
convex solver calls. 

\underline{\itshape Conclusions:} \cref{fig:cactus_obj} demonstrates that 
PEREGRiNN's weighted sum of slacks has a clear benefit over both a uniformly 
weighted sum-of-slacks penalty and a plain feasibility convex program. For 
timeouts of longer than $\approx1.2$ seconds, PEREGRiNN overtakes the other two 
in terms of number of properties proved; even the uniform sum-of-slacks penalty 
considerably outperforms the feasibility convex program at similar timeouts. 
Note that \emph{reversing} the layer-wise weights of PEREGRiNN's penalty 
function incurs a \emph{performance hit}, especially for timeouts of 
$\approx\negthinspace1.2$ seconds. This suggests that driving slacks toward 
shallower layers, where the next neuron is conditioned, is the correct 
heuristic to apply. \cref{fig:histogram_obj} also shows that going from 
feasibility to sum-of-slacks to weighted sum-of-slacks significantly reduces 
the number of test cases that require between 225 and 325 calls to the convex 
solver. This order of comparison shows a concomitant net influx of tests into 
the lowest bin of <25 convex calls; PEREGRiNN has the most test cases in this 
category, with nearly 130 test cases proved in <25 convex solver calls.

\subsubsection{Neuron Conditioning Priority Ablation.} 
\label{ssub:neuron_conditioning_priority_ablation}
In the second ablation experiment, we evaluated the contribution of PEREGRiNN's 
maximum-slack neuron conditioning priority (see 
\cref{sub:max_slack_conditioning_priority}). To that end, we ran variants of 
PEREGRiNN with three different neuron conditioning priorities for the 
search component:
\vspace{-1.25mm}%
\begin{enumerate}
	\item {\itshape ``Maximum slack''}: PEREGRiNN's max-slack neuron 
		conditioning priority;

	\item {\itshape ``Minimum slack''}: This variant conditions the neuron 
		with the smallest slack;

	\item {\itshape ``Random choice''}: This variant conditions on a random 
		indeterminate neuron.
\end{enumerate}
\vspace{-1.25mm}%
The performance of these algorithm variants is shown in 
\cref{fig:cactus_heuristics} and \cref{fig:hist_heuristics}. As in the previous 
ablation experiment, \cref{fig:cactus_heuristics} shows a cactus plot of the 
number of proved cases vs. the timeout, and \cref{fig:hist_heuristics} shows a 
histogram of the number of calls to the convex solver required under each of the 
conditioning priorities.

\underline{\itshape Conclusions:} \cref{fig:cactus_heuristics} shows that 
PEREGRiNN's maximum-slack neuron priority allows it to prove more properties 
for a given timeout than either a random neuron choice priority or the 
minimum-slack choice priority. \cref{fig:hist_heuristics} also shows that the 
benefit of this neuron priority is reflected in a dramatic reduction in the 
number of test cases where 250-350 convex solver calls are required; 
PEREGRiNN's maximum-slack priority has the most test cases requiring <25 calls.
\subsection{Comparison with Other NN Verifiers} 
\label{sub:comparison_with_other_nn_verifiers}
In this experiment, we evaluated the performance of PEREGRiNN with respect to a 
number of state-of-the-art NN verifiers on our adversarial robustness 
testbench, $\mathscr{T}\negthinspace\mathscr{B}$. In particular, we ran the 
following tools on all 300 test cases in $\mathscr{T}\negthinspace\mathscr{B}$: 
Venus \cite{botoeva2020efficient}; Marabou \cite{katz2019marabou}; Neurify 
\cite{wang2018efficient}; and nnenum \cite{BakImprovedGeometricPath2020}. Venus 
was used with \textit{st\_ratio} $=0.4$, \textit{depth\_power} $=4$, 
\textit{offline\_deps} $=$ True, \textit{online\_deps} $=$True, and 
\textit{ideal\_cuts}$=$True; Marabou and Neurify were used with default 
parameters but with \textit{THREADS} $=1$; and nnenum was used with 
\textit{ADVERSARIAL\_SEARCH} turned off. As noted, each algorithm ran inside 
its own single-core VM.

\begin{figure}[t]
    \centering
    \includegraphics[scale = 0.68,trim=0 1.4in 0 1.8in]{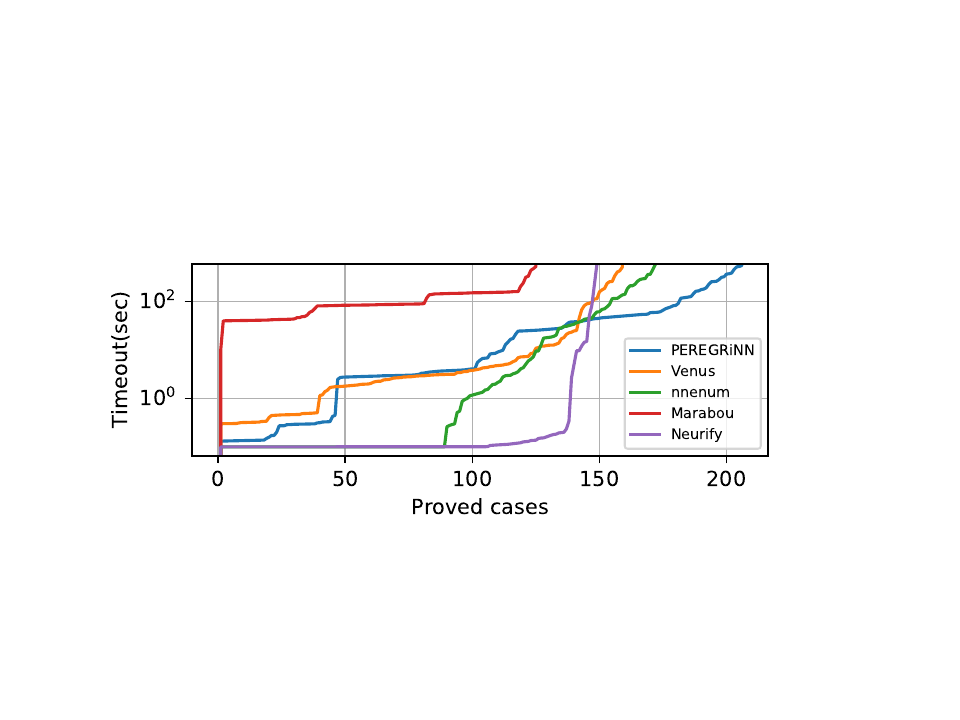}
    \caption{Cactus plot of various solvers on 300-case testbench, $\mathscr{T}\negthinspace\mathscr{B}$}
    \label{fig:solver_comarison_cactus}
\end{figure}

\cref{fig:solver_comarison_cactus} contains a cactus plot showing the results for each of these algorithms, including PEREGRiNN. For a given number of test cases to be proved, \cref{fig:solver_comarison_cactus} depicts the corresponding timeout required for each of the algorithm to prove that many cases. Of all the algorithms, PEREGRiNN was able to prove the most properties within the timeout limit of 300 seconds: PEREGRiNN was able to prove 206 properties; it was followed by nnenum, which proved 172; Venus, which proved 159; Neurify, which proved 149; and Marabou, which proved 125. Marabou consistently performed the worst, proving fewer cases than any other algorithm at every timeout. By contrast, Neurify was able to prove significantly more test cases than any other algorithm for extremely short timeouts, but it failed to prove more than 150 out of 300 test cases across the whole experiment
nnenum performed worse than Neurify on the way to proving 150 test cases, but it fared significantly better than either PEREGRiNN or Venus, which had more or less similar performance below this threshold. However, after $\approx$150 test cases, PEREGRiNN significantly outperformed all other algorithms: as the timeout was increased, PEREGRiNN continued prove additional properties at a rate significantly outpacing its closest competitor in this regime, nnenum.

This data, taken as a whole, suggests that PEREGRiNN suffers from a worse ``best-case'' performance than several other algorithms, especially nnenum and Neurify. However, PEREGRiNN's performance seems to be much more consistent across different test cases. This allows it to prove more properties in aggregate at the expense of being slower on a smaller subset of them. This further suggests that PEREGRiNN is significantly less sensitive to peculiarities of particular test cases on the $\mathscr{T}\negthinspace\mathscr{B}$ testbench. This will likely be a considerable advantage, on average, when faced with verifying unknown networks and properties of this type.

\section{Conclusion}
\label{sec:conclusion}
In this paper, we introduced PEREGRiNN, new tool for formally verifying 
input/output properties for ReLU NNs. PEREGRiNN compares favorably with other 
state-of-the-art NN verifiers, thanks to a number of unique algorithmic 
features. The benefits of these features were established with ablation 
experiments.

\newpage

%
%
%
\bibliographystyle{splncs04}
%
\bibliography{mybib} %

\appendix
\section{Raw Data For Solvers Comparison in section 5.3}
\scriptsize 
\begin{longtable}{|l|l|c|l|l|l|l|l|l|l|l|l|l|}
\hline
\multicolumn{3}{|l|}{} &
  \multicolumn{2}{c|}{\textbf{PEREGRiNN}} &
  \multicolumn{2}{c|}{\textbf{nnenum}} &
  \multicolumn{2}{c|}{\textbf{Venus}} &
  \multicolumn{2}{c|}{\textbf{Neurify}} &
  \multicolumn{2}{c|}{\textbf{Marabou}} \\ \hline
\endhead
\textit{Network} &
  \textit{eps} &
  \textit{image} &
  \textit{result} &
  \textit{time} &
  \textit{result} &
  \textit{time} &
  \textit{result} &
  \textit{time} &
  \textit{result} &
  \textit{time} &
  \textit{result} &
  \textit{time} \\ \hline
MNIST\_FC1 & 0.02 & 1  & unsat & 2.87   & unsat & 45.50  & unsat & 1.78   & unsat & 0.10   & unsat & 147.80 \\ \hline
MNIST\_FC1 & 0.02 & 2  & sat   & 0.29   & sat   & 0.01   & sat   & 0.34   & sat   & 0.01   & sat   & 41.87  \\ \hline
MNIST\_FC1 & 0.02 & 3  & unsat & 16.72  & unsat & 3.23   & unsat & 2.43   & -      & 600.00 & unsat & 335.17 \\ \hline
MNIST\_FC1 & 0.02 & 4  & sat   & 5.55   & sat   & 0.01   & sat   & 1.96   & sat   & 0.01   & sat   & 42.73  \\ \hline
MNIST\_FC1 & 0.02 & 5  & unsat & 3.35   & unsat & 2.29   & unsat & 1.38   & unsat & 0.10   & unsat & 207.41 \\ \hline
MNIST\_FC1 & 0.02 & 6  & sat   & 2.82   & sat   & 1.18   & sat   & 2.04   & -      & 600.00 & sat   & 40.08  \\ \hline
MNIST\_FC1 & 0.02 & 7  & sat   & 9.92   & sat   & 17.66  & sat   & 3.12   & -      & 600.00 & sat   & 63.63  \\ \hline
MNIST\_FC1 & 0.02 & 8  & sat   & 24.98  & sat   & 3.35   & sat   & 3.53   & -      & 600.00 & sat   & 49.43  \\ \hline
MNIST\_FC1 & 0.02 & 9  & unsat & 3.75   & unsat & 1.27   & unsat & 1.15   & unsat & 0.12   & unsat & 150.75 \\ \hline
MNIST\_FC1 & 0.02 & 10 & sat   & 0.13   & sat   & 0.01   & sat   & 0.32   & sat   & 0.01   & sat   & 40.28  \\ \hline
MNIST\_FC1 & 0.02 & 11 & sat   & 0.14   & sat   & 0.01   & sat   & 0.32   & sat   & 0.01   & sat   & 42.47  \\ \hline
MNIST\_FC1 & 0.02 & 12 & sat   & 0.13   & sat   & 0.01   & sat   & 0.34   & sat   & 0.01   & sat   & 44.05  \\ \hline
MNIST\_FC1 & 0.02 & 13 & sat   & 0.14   & sat   & 0.01   & sat   & 0.31   & sat   & 0.01   & sat   & 41.43  \\ \hline
MNIST\_FC1 & 0.02 & 14 & sat   & 0.13   & sat   & 0.01   & sat   & 0.33   & sat   & 0.01   & sat   & 43.50  \\ \hline
MNIST\_FC1 & 0.02 & 15 & sat   & 0.20   & sat   & 0.01   & sat   & 0.40   & sat   & 0.01   & sat   & 48.99  \\ \hline
MNIST\_FC1 & 0.02 & 16 & sat   & 0.13   & sat   & 0.01   & sat   & 2.24   & sat   & 0.01   & sat   & 43.07  \\ \hline
MNIST\_FC1 & 0.02 & 17 & sat   & 2.48   & sat   & 0.01   & sat   & 1.75   & sat   & 0.01   & sat   & 43.16  \\ \hline
MNIST\_FC1 & 0.02 & 18 & sat   & 3.02   & sat   & 1.49   & sat   & 1.95   & -      & 600.00 & sat   & 43.34  \\ \hline
MNIST\_FC1 & 0.02 & 19 & unsat & 2.85   & unsat & 3.01   & unsat & 1.19   & unsat & 0.10   & unsat & 445.00 \\ \hline
MNIST\_FC1 & 0.02 & 20 & sat   & 2.71   & sat   & 0.30   & sat   & 2.54   & -      & 600.00 & sat   & 52.60  \\ \hline
MNIST\_FC1 & 0.02 & 21 & sat   & 2.90   & sat   & 0.01   & sat   & 1.84   & sat   & 0.01   & sat   & 42.62  \\ \hline
MNIST\_FC1 & 0.02 & 22 & sat   & 24.66  & sat   & 44.35  & sat   & 10.95  & -      & 600.00 & sat   & 90.05  \\ \hline
MNIST\_FC1 & 0.02 & 23 & sat   & 0.14   & sat   & 0.01   & sat   & 0.32   & sat   & 0.01   & sat   & 46.66  \\ \hline
MNIST\_FC1 & 0.02 & 24 & sat   & 0.13   & sat   & 0.01   & sat   & 0.32   & sat   & 0.01   & sat   & 40.42  \\ \hline
MNIST\_FC1 & 0.02 & 25 & unsat & 3.15   & unsat & 5.97   & unsat & 1.48   & unsat & 0.10   & -      & 600.00 \\ \hline
MNIST\_FC1 & 0.02 & 26 & sat   & 0.14   & sat   & 0.01   & sat   & 2.41   & sat   & 0.01   & sat   & 40.44  \\ \hline
MNIST\_FC1 & 0.02 & 27 & sat   & 25.35  & sat   & 1.23   & sat   & 1.96   & -      & 600.00 & sat   & 148.23 \\ \hline
MNIST\_FC1 & 0.02 & 28 & sat   & 21.22  & sat   & 0.03   & sat   & 0.30   & -      & 600.00 & sat   & 40.55  \\ \hline
MNIST\_FC1 & 0.02 & 29 & sat   & 0.14   & sat   & 0.01   & sat   & 0.30   & sat   & 0.01   & sat   & 43.06  \\ \hline
MNIST\_FC1 & 0.02 & 30 & sat   & 0.14   & sat   & 0.01   & sat   & 0.32   & sat   & 0.17   & sat   & 41.28  \\ \hline
MNIST\_FC1 & 0.02 & 31 & sat   & 0.17   & sat   & 0.01   & -      & 600.09 & sat   & 0.01   & sat   & 41.57  \\ \hline
MNIST\_FC1 & 0.02 & 32 & sat   & 117.93 & -      & 600.17 & sat   & 0.30   & -      & 599.32 & -      & 600.17 \\ \hline
MNIST\_FC1 & 0.02 & 33 & sat   & 0.13   & sat   & 0.01   & sat   & 1.69   & sat   & 0.01   & sat   & 41.25  \\ \hline
MNIST\_FC1 & 0.02 & 34 & sat   & 0.13   & sat   & 0.01   & sat   & 1.76   & sat   & 0.01   & sat   & 40.99  \\ \hline
MNIST\_FC1 & 0.02 & 35 & sat   & 2.99   & sat   & 0.51   & sat   & 0.30   & sat   & 0.19   & sat   & 60.32  \\ \hline
MNIST\_FC1 & 0.02 & 36 & sat   & 0.14   & sat   & 0.01   & sat   & 4.73   & sat   & 0.01   & sat   & 42.33  \\ \hline
MNIST\_FC1 & 0.02 & 37 & sat   & 6.19   & sat   & 28.06  & sat   & 0.32   & -      & 600.00 & sat   & 82.00  \\ \hline
MNIST\_FC1 & 0.02 & 38 & sat   & 0.17   & sat   & 2.15   & unsat & 1.89   & -      & 600.00 & sat   & 41.06  \\ \hline
MNIST\_FC1 & 0.02 & 39 & unsat & 66.62  & unsat & 61.86  & sat   & 0.30   & sat   & 0.13   & -      & 600.17 \\ \hline
MNIST\_FC1 & 0.02 & 40 & sat   & 0.14   & sat   & 0.01   & sat   & 4.07   & sat   & 0.01   & sat   & 41.09  \\ \hline
MNIST\_FC1 & 0.02 & 41 & sat   & 25.52  & sat   & 28.66  & sat   & 0.30   & -      & 600.00 & sat   & 319.10 \\ \hline
MNIST\_FC1 & 0.02 & 42 & sat   & 0.15   & sat   & 0.01   & sat   & 2.27   & sat   & 0.01   & sat   & 40.80  \\ \hline
MNIST\_FC1 & 0.02 & 43 & sat   & 8.45   & sat   & 0.26   & sat   & 7.24   & -      & 600.00 & sat   & 40.76  \\ \hline
MNIST\_FC1 & 0.02 & 44 & sat   & 122.21 & -      & 600.03 & sat   & 1.80   & -      & 600.00 & -      & 600.17 \\ \hline
MNIST\_FC1 & 0.02 & 45 & sat   & 0.16   & sat   & 0.01   & sat   & 1.77   & sat   & 0.01   & sat   & 41.44  \\ \hline
MNIST\_FC1 & 0.02 & 46 & sat   & 12.88  & sat   & 0.01   & sat   & 7.38   & sat   & 0.01   & sat   & 40.50  \\ \hline
MNIST\_FC1 & 0.02 & 47 & sat   & 6.70   & sat   & 141.47 & sat   & 0.30   & -      & 600.00 & sat   & 72.04  \\ \hline
MNIST\_FC1 & 0.02 & 48 & sat   & 0.13   & sat   & 0.01   & sat   & 0.32   & sat   & 0.01   & sat   & 46.75  \\ \hline
MNIST\_FC1 & 0.02 & 49 & sat   & 0.13   & sat   & 0.01   & sat   & 0.32   & sat   & 0.01   & sat   & 40.79  \\ \hline
MNIST\_FC1 & 0.02 & 50 & sat   & 0.13   & sat   & 0.01   & sat   & 1.84   & sat   & 0.01   & sat   & 40.46  \\ \hline
MNIST\_FC2 & 0.02 & 1  & sat   & 0.30   & sat   & 0.03   & sat   & 0.46   & sat   & 0.01   & sat   & 83.14  \\ \hline
MNIST\_FC2 & 0.02 & 2  & sat   & 0.27   & sat   & 0.02   & sat   & 0.46   & sat   & 0.01   & sat   & 81.80  \\ \hline
MNIST\_FC2 & 0.02 & 3  & sat   & 46.46  & -      & 600.21 & -      & 600.10 & -      & 599.09 & -      & 600.00 \\ \hline
MNIST\_FC2 & 0.02 & 4  & sat   & 0.43   & sat   & 0.02   & sat   & 0.46   & sat   & 0.01   & sat   & 84.07  \\ \hline
MNIST\_FC2 & 0.02 & 5  & sat   & 0.29   & sat   & 0.02   & sat   & 0.49   & sat   & 0.01   & sat   & 85.36  \\ \hline
MNIST\_FC2 & 0.02 & 6  & sat   & 38.37  & sat   & 6.48   & sat   & 7.21   & sat   & 9.69   & sat   & 85.53  \\ \hline
MNIST\_FC2 & 0.02 & 7  & sat   & 0.31   & sat   & 0.02   & sat   & 0.46   & sat   & 0.01   & sat   & 84.60  \\ \hline
MNIST\_FC2 & 0.02 & 8  & sat   & 377.13 & -      & 600.03 & -      & 600.10 & -      & 599.60 & -      & 600.17 \\ \hline
MNIST\_FC2 & 0.02 & 9  & sat   & 28.20  & sat   & 3.58   & sat   & 5.76   & sat   & 83.70  & sat   & 84.11  \\ \hline
MNIST\_FC2 & 0.02 & 10 & sat   & 0.30   & sat   & 0.02   & sat   & 0.50   & sat   & 0.01   & sat   & 82.22  \\ \hline
MNIST\_FC2 & 0.02 & 11 & sat   & 27.98  & sat   & 0.02   & sat   & 3.10   & sat   & 0.01   & sat   & 83.40  \\ \hline
MNIST\_FC2 & 0.02 & 12 & sat   & 0.33   & sat   & 0.02   & sat   & 0.49   & sat   & 0.01   & sat   & 82.91  \\ \hline
MNIST\_FC2 & 0.02 & 13 & sat   & 0.33   & sat   & 0.02   & sat   & 0.50   & sat   & 0.01   & sat   & 82.98  \\ \hline
MNIST\_FC2 & 0.02 & 14 & sat   & 0.33   & sat   & 0.02   & sat   & 0.49   & sat   & 0.01   & sat   & 84.93  \\ \hline
MNIST\_FC2 & 0.02 & 15 & sat   & 27.18  & sat   & 0.02   & sat   & 3.10   & sat   & 0.01   & sat   & 84.90  \\ \hline
MNIST\_FC2 & 0.02 & 16 & sat   & 0.33   & sat   & 0.02   & sat   & 0.51   & sat   & 0.01   & sat   & 84.64  \\ \hline
MNIST\_FC2 & 0.02 & 17 & sat   & 39.31  & sat   & 9.57   & sat   & 67.99  & sat   & 9.70   & sat   & 89.17  \\ \hline
MNIST\_FC2 & 0.02 & 18 & sat   & 0.30   & sat   & 0.02   & sat   & 0.49   & sat   & 0.01   & sat   & 84.23  \\ \hline
MNIST\_FC2 & 0.02 & 19 & unsat & 8.56   & -      & 600.12 & unsat & 12.02  & sat   & 0.13   & -      & 600.00 \\ \hline
MNIST\_FC2 & 0.02 & 20 & sat   & 27.23  & sat   & 0.02   & sat   & 3.17   & sat   & 0.01   & sat   & 88.61  \\ \hline
MNIST\_FC2 & 0.02 & 21 & sat   & 42.08  & sat   & 296.30 & sat   & 97.34  & -      & 599.87 & -      & 600.00 \\ \hline
MNIST\_FC2 & 0.02 & 22 & sat   & 26.89  & sat   & 0.03   & sat   & 3.41   & sat   & 0.01   & sat   & 91.16  \\ \hline
MNIST\_FC2 & 0.02 & 23 & sat   & 0.30   & sat   & 0.02   & sat   & 0.46   & sat   & 0.01   & sat   & 87.52  \\ \hline
MNIST\_FC2 & 0.02 & 24 & sat   & 0.29   & sat   & 0.02   & sat   & 0.46   & sat   & 0.01   & sat   & 87.49  \\ \hline
MNIST\_FC2 & 0.02 & 25 & sat   & 40.33  & sat   & 9.58   & sat   & 13.91  & sat   & 14.46  & sat   & 89.15  \\ \hline
MNIST\_FC2 & 0.02 & 26 & sat   & 0.30   & sat   & 0.02   & sat   & 3.61   & sat   & 0.01   & sat   & 84.77  \\ \hline
MNIST\_FC2 & 0.02 & 27 & sat   & 38.32  & sat   & 0.03   & sat   & 0.47   & sat   & 0.01   & sat   & 89.09  \\ \hline
MNIST\_FC2 & 0.02 & 28 & sat   & 0.27   & sat   & 0.02   & sat   & 2.87   & sat   & 0.01   & sat   & 86.84  \\ \hline
MNIST\_FC2 & 0.02 & 29 & sat   & 9.10   & sat   & 0.02   & -      & 600.10 & sat   & 0.01   & sat   & 85.41  \\ \hline
MNIST\_FC2 & 0.02 & 30 & sat   & 60.42  & sat   & 363.88 & sat   & 2.94   & -      & 600.12 & -      & 600.00 \\ \hline
MNIST\_FC2 & 0.02 & 31 & sat   & 24.65  & sat   & 0.03   & -      & 600.10 & sat   & 0.01   & sat   & 86.74  \\ \hline
MNIST\_FC2 & 0.02 & 32 & sat   & 466.61 & -      & 600.29 & sat   & 2.78   & -      & 599.54 & -      & 600.00 \\ \hline
MNIST\_FC2 & 0.02 & 33 & sat   & 24.85  & sat   & 0.02   & sat   & 5.07   & sat   & 0.01   & sat   & 88.35  \\ \hline
MNIST\_FC2 & 0.02 & 34 & sat   & 33.97  & sat   & 4.32   & sat   & 0.45   & sat   & 5.02   & sat   & 89.53  \\ \hline
MNIST\_FC2 & 0.02 & 35 & sat   & 0.29   & sat   & 0.02   & sat   & 0.47   & sat   & 0.01   & sat   & 88.17  \\ \hline
MNIST\_FC2 & 0.02 & 36 & sat   & 0.29   & sat   & 0.02   & sat   & 2.99   & sat   & 0.01   & sat   & 87.45  \\ \hline
MNIST\_FC2 & 0.02 & 37 & sat   & 0.30   & sat   & 0.02   & sat   & 21.15  & sat   & 0.01   & sat   & 87.95  \\ \hline
MNIST\_FC2 & 0.02 & 38 & sat   & 39.76  & sat   & 17.93  & -      & 600.10 & -      & 599.70 & sat   & 89.60  \\ \hline
MNIST\_FC2 & 0.02 & 39 & sat   & 370.71 & -      & 600.10 & sat   & 5.36   & -      & 600.17 & -      & 600.00 \\ \hline
MNIST\_FC2 & 0.02 & 40 & sat   & 25.83  & sat   & 7.30   & -      & 600.10 & sat   & 15.11  & sat   & 88.98  \\ \hline
MNIST\_FC2 & 0.02 & 41 & -      & 600.00 & -      & 600.11 & sat   & 82.48  & -      & 599.27 & -      & 600.00 \\ \hline
MNIST\_FC2 & 0.02 & 42 & sat   & 38.30  & sat   & 30.97  & sat   & 0.44   & -      & 599.87 & -      & 609.27 \\ \hline
MNIST\_FC2 & 0.02 & 43 & sat   & 0.27   & sat   & 0.02   & unsat & 4.55   & sat   & 0.01   & sat   & 84.04  \\ \hline
MNIST\_FC2 & 0.02 & 44 & unsat & 327.70 & unsat & 212.93 & sat   & 2.67   & sat   & 0.14   & -      & 600.00 \\ \hline
MNIST\_FC2 & 0.02 & 45 & sat   & 24.97  & sat   & 0.03   & sat   & 4.27   & sat   & 0.01   & sat   & 86.14  \\ \hline
MNIST\_FC2 & 0.02 & 46 & sat   & 26.63  & sat   & 4.09   & -      & 600.11 & -      & 598.87 & sat   & 85.63  \\ \hline
MNIST\_FC2 & 0.02 & 47 & sat   & 227.45 & -      & 600.13 & sat   & 3.15   & -      & 599.90 & -      & 610.70 \\ \hline
MNIST\_FC2 & 0.02 & 48 & sat   & 26.07  & sat   & 0.03   & sat   & 0.45   & sat   & 0.01   & sat   & 86.79  \\ \hline
MNIST\_FC2 & 0.02 & 49 & sat   & 0.32   & sat   & 0.02   & sat   & 0.45   & sat   & 0.01   & sat   & 81.58  \\ \hline
MNIST\_FC2 & 0.02 & 50 & sat   & 0.29   & sat   & 0.02   & sat   & 3.05   & sat   & 0.01   & sat   & 85.61  \\ \hline
MNIST\_FC3 & 0.02 & 1  & sat   & 54.33  & sat   & 0.04   & -      & 600.11 & sat   & 0.01   & sat   & 160.98 \\ \hline
MNIST\_FC3 & 0.02 & 2  & sat   & 27.50  & sat   & 0.03   & sat   & 3.61   & sat   & 0.01   & sat   & 153.05 \\ \hline
MNIST\_FC3 & 0.02 & 3  & -      & 600.00 & -      & 600.18 & -      & 600.10 & -      & 600.05 & -      & 600.00 \\ \hline
MNIST\_FC3 & 0.02 & 4  & sat   & 48.95  & sat   & 0.04   & -      & 600.04 & sat   & 0.01   & sat   & 154.26 \\ \hline
MNIST\_FC3 & 0.02 & 5  & sat   & 54.60  & sat   & 0.04   & -      & 600.11 & sat   & 0.01   & sat   & 156.33 \\ \hline
MNIST\_FC3 & 0.02 & 6  & sat   & 263.79 & sat   & 36.70  & -      & 600.10 & -      & 600.09 & sat   & 158.42 \\ \hline
MNIST\_FC3 & 0.02 & 7  & -      & 600.00 & -      & 600.18 & -      & 600.10 & -      & 601.68 & -      & 600.00 \\ \hline
MNIST\_FC3 & 0.02 & 8  & sat   & 60.05  & sat   & 12.65  & -      & 600.10 & sat   & 51.00  & sat   & 146.29 \\ \hline
MNIST\_FC3 & 0.02 & 9  & unsat & 82.26  & unsat & 366.96 & unsat & 8.61   & unsat & 0.18   & -      & 600.00 \\ \hline
MNIST\_FC3 & 0.02 & 10 & sat   & 51.89  & sat   & 0.04   & -      & 600.11 & sat   & 0.01   & sat   & 146.48 \\ \hline
MNIST\_FC3 & 0.02 & 11 & sat   & 47.90  & sat   & 0.04   & sat   & 259.19 & sat   & 0.01   & sat   & 155.87 \\ \hline
MNIST\_FC3 & 0.02 & 12 & sat   & 62.09  & sat   & 0.04   & -      & 600.11 & sat   & 0.01   & sat   & 155.02 \\ \hline
MNIST\_FC3 & 0.02 & 13 & -      & 600.00 & -      & 600.05 & -      & 600.10 & -      & 599.40 & -      & 600.00 \\ \hline
MNIST\_FC3 & 0.02 & 14 & sat   & 54.23  & sat   & 0.04   & -      & 600.10 & sat   & 0.01   & sat   & 154.00 \\ \hline
MNIST\_FC3 & 0.02 & 15 & sat   & 59.44  & sat   & 0.04   & -      & 600.10 & sat   & 0.01   & sat   & 151.65 \\ \hline
MNIST\_FC3 & 0.02 & 16 & sat   & 54.57  & sat   & 0.04   & -      & 600.10 & sat   & 0.01   & sat   & 149.86 \\ \hline
MNIST\_FC3 & 0.02 & 17 & -      & 600.00 & -      & 600.45 & -      & 600.10 & -      & 601.82 & -      & 600.00 \\ \hline
MNIST\_FC3 & 0.02 & 18 & sat   & 55.33  & sat   & 0.03   & -      & 600.10 & sat   & 0.01   & sat   & 157.17 \\ \hline
MNIST\_FC3 & 0.02 & 19 & -      & 600.00 & -      & 600.12 & -      & 600.10 & -      & 599.58 & -      & 600.00 \\ \hline
MNIST\_FC3 & 0.02 & 20 & sat   & 53.02  & sat   & 0.03   & -      & 600.11 & sat   & 0.01   & sat   & 161.45 \\ \hline
MNIST\_FC3 & 0.02 & 21 & sat   & 91.87  & sat   & 19.24  & -      & 600.11 & sat   & 12.14  & sat   & 153.13 \\ \hline
MNIST\_FC3 & 0.02 & 22 & sat   & 59.33  & sat   & 0.04   & -      & 600.11 & sat   & 0.01   & sat   & 153.29 \\ \hline
MNIST\_FC3 & 0.02 & 23 & sat   & 51.93  & sat   & 0.04   & -      & 600.10 & sat   & 0.01   & sat   & 154.05 \\ \hline
MNIST\_FC3 & 0.02 & 24 & sat   & 49.11  & sat   & 0.04   & -      & 600.10 & sat   & 0.01   & sat   & 156.44 \\ \hline
MNIST\_FC3 & 0.02 & 25 & -      & 600.00 & -      & 600.08 & -      & 600.10 & -      & 599.40 & -      & 600.00 \\ \hline
MNIST\_FC3 & 0.02 & 26 & sat   & 0.45   & sat   & 0.03   & -      & 600.11 & sat   & 0.01   & sat   & 144.99 \\ \hline
MNIST\_FC3 & 0.02 & 27 & sat   & 48.22  & sat   & 0.04   & sat   & 340.83 & sat   & 0.01   & sat   & 159.00 \\ \hline
MNIST\_FC3 & 0.02 & 28 & sat   & 45.62  & sat   & 0.03   & -      & 600.10 & sat   & 0.01   & sat   & 152.76 \\ \hline
MNIST\_FC3 & 0.02 & 29 & sat   & 47.22  & sat   & 0.04   & -      & 600.10 & sat   & 0.01   & sat   & 152.33 \\ \hline
MNIST\_FC3 & 0.02 & 30 & -      & 600.00 & -      & 600.46 & sat   & 6.06   & -      & 600.16 & -      & 600.00 \\ \hline
MNIST\_FC3 & 0.02 & 31 & sat   & 26.37  & sat   & 0.03   & -      & 600.10 & sat   & 0.01   & sat   & 156.86 \\ \hline
MNIST\_FC3 & 0.02 & 32 & -      & 600.00 & -      & 603.10 & sat   & 517.73 & -      & 601.30 & -      & 600.00 \\ \hline
MNIST\_FC3 & 0.02 & 33 & sat   & 44.60  & sat   & 0.04   & unsat & 259.14 & sat   & 0.01   & sat   & 149.50 \\ \hline
MNIST\_FC3 & 0.02 & 34 & unsat & 257.10 & -      & 600.62 & -      & 600.10 & unsat & 0.17   & -      & 600.00 \\ \hline
MNIST\_FC3 & 0.02 & 35 & sat   & 50.51  & sat   & 0.04   & sat   & 8.51   & sat   & 0.01   & sat   & 149.53 \\ \hline
MNIST\_FC3 & 0.02 & 36 & sat   & 41.34  & sat   & 0.04   & -      & 600.10 & sat   & 0.01   & sat   & 148.25 \\ \hline
MNIST\_FC3 & 0.02 & 37 & -      & 600.00 & -      & 600.75 & -      & 600.10 & sat   & 0.18   & -      & 600.00 \\ \hline
MNIST\_FC3 & 0.02 & 38 & -      & 600.00 & -      & 600.41 & -      & 600.10 & -      & 599.41 & -      & 600.00 \\ \hline
MNIST\_FC3 & 0.02 & 39 & sat   & 534.05 & sat   & 587.99 & -      & 600.10 & -      & 599.38 & -      & 600.00 \\ \hline
MNIST\_FC3 & 0.02 & 40 & sat   & 51.64  & sat   & 0.04   & -      & 600.11 & sat   & 0.01   & sat   & 148.09 \\ \hline
MNIST\_FC3 & 0.02 & 41 & -      & 600.00 & -      & 602.60 & -      & 600.11 & -      & 601.68 & -      & 600.00 \\ \hline
MNIST\_FC3 & 0.02 & 42 & -      & 600.00 & -      & 600.71 & -      & 600.11 & -      & 599.34 & -      & 600.00 \\ \hline
MNIST\_FC3 & 0.02 & 43 & sat   & 50.81  & sat   & 0.04   & -      & 600.11 & sat   & 0.01   & sat   & 144.15 \\ \hline
MNIST\_FC3 & 0.02 & 44 & -      & 600.00 & -      & 600.27 & -      & 600.11 & -      & 600.90 & -      & 600.00 \\ \hline
MNIST\_FC3 & 0.02 & 45 & sat   & 52.56  & sat   & 18.55  & -      & 600.10 & sat   & 2.65   & sat   & 145.92 \\ \hline
MNIST\_FC3 & 0.02 & 46 & sat   & 146.42 & sat   & 116.21 & -      & 600.10 & -      & 599.56 & -      & 600.00 \\ \hline
MNIST\_FC3 & 0.02 & 47 & -      & 600.00 & -      & 602.25 & -      & 600.10 & -      & 607.98 & -      & 600.00 \\ \hline
MNIST\_FC3 & 0.02 & 48 & sat   & 44.44  & sat   & 0.04   & sat   & 3.94   & sat   & 0.01   & sat   & 140.50 \\ \hline
MNIST\_FC3 & 0.02 & 49 & sat   & 25.71  & sat   & 0.03   & -      & 600.09 & sat   & 0.01   & sat   & 138.45 \\ \hline
MNIST\_FC3 & 0.02 & 50 & sat   & 46.96  & sat   & 0.03   & -      & 600.11 & sat   & 0.01   & sat   & 145.12 \\ \hline
MNIST\_FC1 & 0.05 & 1  & unsat & 3.99   & unsat & 0.06   & unsat & 3.88   & unsat & 0.16   & -      & 600.00 \\ \hline
MNIST\_FC1 & 0.05 & 2  & unsat & 2.76   & unsat & 78.65  & unsat & 1.72   & unsat & 0.11   & -      & 600.00 \\ \hline
MNIST\_FC1 & 0.05 & 3  & unsat & 4.10   & -      & 600.11 & unsat & 4.38   & -      & 600.01 & -      & 600.00 \\ \hline
MNIST\_FC1 & 0.05 & 4  & unsat & 3.43   & -      & 600.01 & unsat & 23.41  & -      & 600.29 & -      & 600.00 \\ \hline
MNIST\_FC1 & 0.05 & 5  & unsat & 3.05   & unsat & 1.13   & unsat & 3.16   & unsat & 0.09   & -      & 600.00 \\ \hline
MNIST\_FC1 & 0.05 & 6  & unsat & 3.64   & unsat & 462.01 & unsat & 2.90   & unsat & 0.14   & -      & 600.00 \\ \hline
MNIST\_FC1 & 0.05 & 7  & unsat & 2.98   & unsat & 127.57 & unsat & 3.14   & unsat & 0.10   & -      & 600.00 \\ \hline
MNIST\_FC1 & 0.05 & 8  & unsat & 3.73   & unsat & 61.14  & unsat & 3.79   & unsat & 0.08   & -      & 600.00 \\ \hline
MNIST\_FC1 & 0.05 & 9  & unsat & 2.97   & unsat & 2.77   & unsat & 2.81   & unsat & 0.12   & unsat & 82.44  \\ \hline
MNIST\_FC1 & 0.05 & 10 & unsat & 2.80   & unsat & 33.17  & unsat & 2.16   & unsat & 0.10   & -      & 628.60 \\ \hline
MNIST\_FC1 & 0.05 & 11 & sat   & 30.31  & sat   & 39.27  & sat   & 4.90   & -      & 149.19 & sat   & 481.15 \\ \hline
MNIST\_FC1 & 0.05 & 12 & unsat & 2.95   & unsat & 0.95   & unsat & 1.90   & unsat & 0.09   & -      & 600.00 \\ \hline
MNIST\_FC1 & 0.05 & 13 & unsat & 2.95   & unsat & 43.18  & unsat & 2.68   & unsat & 0.10   & -      & 600.00 \\ \hline
MNIST\_FC1 & 0.05 & 14 & unsat & 17.37  & -      & 600.02 & unsat & 45.11  & -      & 600.62 & -      & 600.00 \\ \hline
MNIST\_FC1 & 0.05 & 15 & unsat & 3.70   & -      & 600.06 & unsat & 92.66  & -      & 599.81 & -      & 600.00 \\ \hline
MNIST\_FC1 & 0.05 & 16 & sat   & 166.52 & -      & 600.05 & sat   & 114.13 & -      & 600.62 & -      & 600.00 \\ \hline
MNIST\_FC1 & 0.05 & 17 & unsat & 3.50   & unsat & 116.84 & unsat & 1.82   & -      & 599.81 & -      & 600.00 \\ \hline
MNIST\_FC1 & 0.05 & 18 & unsat & 2.89   & unsat & 1.30   & unsat & 3.01   & unsat & 0.11   & -      & 600.00 \\ \hline
MNIST\_FC1 & 0.05 & 19 & unsat & 2.81   & unsat & 0.05   & unsat & 1.99   & unsat & 0.09   & -      & 600.00 \\ \hline
MNIST\_FC1 & 0.05 & 20 & unsat & 3.82   & unsat & 38.17  & unsat & 7.35   & unsat & 0.08   & -      & 600.00 \\ \hline
MNIST\_FC1 & 0.05 & 21 & unsat & 3.58   & unsat & 30.79  & unsat & 2.78   & unsat & 0.09   & -      & 600.00 \\ \hline
MNIST\_FC1 & 0.05 & 22 & unsat & 3.73   & unsat & 32.08  & unsat & 2.82   & unsat & 0.08   & -      & 600.00 \\ \hline
MNIST\_FC1 & 0.05 & 23 & sat   & 83.75  & sat   & 299.29 & sat   & 22.71  & -      & 600.62 & sat   & 532.14 \\ \hline
MNIST\_FC1 & 0.05 & 24 & unsat & 2.87   & -      & 600.02 & unsat & 12.66  & unsat & 0.11   & -      & 600.00 \\ \hline
MNIST\_FC1 & 0.05 & 25 & unsat & 3.84   & unsat & 1.34   & unsat & 4.89   & unsat & 0.09   & unsat & 10.04  \\ \hline
MNIST\_FC1 & 0.05 & 26 & unsat & 2.77   & unsat & 71.18  & unsat & 3.15   & -      & 600.62 & unsat & 118.04 \\ \hline
MNIST\_FC1 & 0.05 & 27 & unsat & 3.87   & unsat & 1.74   & unsat & 4.78   & unsat & 0.11   & -      & 600.00 \\ \hline
MNIST\_FC1 & 0.05 & 28 & unsat & 3.33   & unsat & 1.00   & sat   & 6.95   & unsat & 0.09   & -      & 600.00 \\ \hline
MNIST\_FC1 & 0.05 & 29 & sat   & 125.23 & sat   & 186.32 & unsat & 24.82  & -      & 600.62 & sat   & 241.73 \\ \hline
MNIST\_FC1 & 0.05 & 30 & unsat & 3.72   & -      & 600.01 & unsat & 2.94   & -      & 600.06 & -      & 600.00 \\ \hline
MNIST\_FC1 & 0.05 & 31 & unsat & 2.86   & unsat & 265.43 & unsat & 17.11  & -      & 599.90 & -      & 600.00 \\ \hline
MNIST\_FC1 & 0.05 & 32 & unsat & 4.08   & unsat & 2.96   & unsat & 3.80   & -      & 600.60 & -      & 600.00 \\ \hline
MNIST\_FC1 & 0.05 & 33 & unsat & 2.83   & unsat & 213.16 & unsat & 113.68 & unsat & 0.11   & -      & 600.00 \\ \hline
MNIST\_FC1 & 0.05 & 34 & sat   & 390.85 & unsat & 18.37  & unsat & 2.71   & -      & 601.73 & -      & 600.00 \\ \hline
MNIST\_FC1 & 0.05 & 35 & unsat & 3.01   & unsat & 115.65 & unsat & 2.27   & unsat & 0.09   & -      & 600.00 \\ \hline
MNIST\_FC1 & 0.05 & 36 & unsat & 2.87   & unsat & 2.98   & unsat & 3.07   & unsat & 0.09   & -      & 600.00 \\ \hline
MNIST\_FC1 & 0.05 & 37 & unsat & 2.96   & unsat & 287.02 & unsat & 12.31  & unsat & 0.12   & -      & 600.00 \\ \hline
MNIST\_FC1 & 0.05 & 38 & unsat & 3.69   & -      & 600.15 & unsat & 12.58  & -      & 600.45 & -      & 600.00 \\ \hline
MNIST\_FC1 & 0.05 & 39 & unsat & 3.05   & unsat & 1.95   & unsat & 2.93   & unsat & 0.11   & -      & 600.00 \\ \hline
MNIST\_FC1 & 0.05 & 40 & unsat & 2.91   & -      & 600.02 & unsat & 4.60   & -      & 600.81 & -      & 600.00 \\ \hline
MNIST\_FC1 & 0.05 & 41 & unsat & 4.04   & -      & 600.37 & unsat & 25.38  & -      & 601.91 & -      & 600.00 \\ \hline
MNIST\_FC1 & 0.05 & 42 & unsat & 3.90   & unsat & 5.06   & unsat & 5.13   & unsat & 0.12   & -      & 600.00 \\ \hline
MNIST\_FC1 & 0.05 & 43 & unsat & 3.09   & unsat & 54.09  & unsat & 12.05  & unsat & 0.10   & -      & 600.00 \\ \hline
MNIST\_FC1 & 0.05 & 44 & unsat & 3.67   & unsat & 137.82 & unsat & 13.24  & unsat & 0.11   & -      & 600.00 \\ \hline
MNIST\_FC1 & 0.05 & 45 & unsat & 3.01   & unsat & 1.97   & unsat & 4.34   & -      & 599.82 & -      & 600.00 \\ \hline
MNIST\_FC1 & 0.05 & 46 & unsat & 2.98   & unsat & 45.14  & unsat & 12.70  & unsat & 0.11   & -      & 600.00 \\ \hline
MNIST\_FC1 & 0.05 & 47 & unsat & 2.98   & unsat & 1.53   & unsat & 11.23  & -      & 599.88 & -      & 600.00 \\ \hline
MNIST\_FC1 & 0.05 & 48 & unsat & 3.53   & unsat & 69.15  & unsat & 2.47   & unsat & 0.11   & -      & 600.00 \\ \hline
MNIST\_FC1 & 0.05 & 49 & unsat & 2.93   & unsat & 0.55   & unsat & 2.64   & unsat & 0.10   & -      & 600.00 \\ \hline
MNIST\_FC1 & 0.05 & 50 & unsat & 2.84   & unsat & 0.87   & unsat & 3.41   & unsat & 0.09   & -      & 600.00 \\ \hline
MNIST\_FC2 & 0.05 & 1  & -      & 600.00 & -      & 600.12 & -      & 600.10 & -      & 600.67 & -      & 600.00 \\ \hline
MNIST\_FC2 & 0.05 & 2  & -      & 600.00 & -      & 600.54 & -      & 600.11 & -      & 599.21 & -      & 600.00 \\ \hline
MNIST\_FC2 & 0.05 & 3  & -      & 600.00 & -      & 602.04 & -      & 600.10 & -      & 600.12 & -      & 600.00 \\ \hline
MNIST\_FC2 & 0.05 & 4  & sat   & 524.35 & -      & 600.16 & -      & 600.10 & -      & 599.98 & -      & 600.00 \\ \hline
MNIST\_FC2 & 0.05 & 5  & -      & 600.00 & -      & 600.48 & -      & 600.11 & -      & 600.17 & -      & 600.00 \\ \hline
MNIST\_FC2 & 0.05 & 6  & -      & 600.00 & -      & 600.54 & -      & 600.10 & -      & 601.22 & -      & 600.00 \\ \hline
MNIST\_FC2 & 0.05 & 7  & -      & 600.00 & -      & 600.87 & -      & 600.10 & -      & 599.89 & -      & 600.00 \\ \hline
MNIST\_FC2 & 0.05 & 8  & unsat & 119.51 & -      & 600.22 & -      & 600.11 & -      & 602.95 & -      & 600.00 \\ \hline
MNIST\_FC2 & 0.05 & 9  & -      & 600.00 & -      & 600.68 & -      & 600.10 & -      & 599.42 & -      & 600.00 \\ \hline
MNIST\_FC2 & 0.05 & 10 & -      & 600.00 & -      & 600.01 & -      & 600.11 & -      & 601.33 & -      & 600.00 \\ \hline
MNIST\_FC2 & 0.05 & 11 & -      & 600.00 & -      & 600.30 & -      & 600.10 & -      & 599.55 & -      & 600.00 \\ \hline
MNIST\_FC2 & 0.05 & 12 & -      & 600.00 & -      & 600.02 & -      & 600.11 & -      & 599.43 & -      & 600.00 \\ \hline
MNIST\_FC2 & 0.05 & 13 & -      & 600.00 & -      & 600.97 & -      & 600.10 & -      & 601.35 & -      & 600.00 \\ \hline
MNIST\_FC2 & 0.05 & 14 & -      & 600.00 & -      & 600.31 & -      & 600.11 & -      & 600.01 & -      & 600.00 \\ \hline
MNIST\_FC2 & 0.05 & 15 & unsat & 181.77 & -      & 600.17 & -      & 600.11 & -      & 602.49 & -      & 600.00 \\ \hline
MNIST\_FC2 & 0.05 & 16 & -      & 600.00 & -      & 601.37 & -      & 600.10 & -      & 600.66 & -      & 600.00 \\ \hline
MNIST\_FC2 & 0.05 & 17 & -      & 600.00 & -      & 601.21 & -      & 600.10 & -      & 601.30 & -      & 600.00 \\ \hline
MNIST\_FC2 & 0.05 & 18 & -      & 600.00 & -      & 600.13 & -      & 600.11 & -      & 599.20 & -      & 600.00 \\ \hline
MNIST\_FC2 & 0.05 & 19 & unsat & 6.89   & -      & 600.17 & -      & 600.11 & unsat & 0.20   & -      & 600.00 \\ \hline
MNIST\_FC2 & 0.05 & 20 & -      & 600.00 & -      & 600.43 & -      & 600.10 & -      & 600.50 & -      & 600.00 \\ \hline
MNIST\_FC2 & 0.05 & 21 & -      & 600.00 & -      & 601.10 & -      & 600.10 & -      & 600.35 & -      & 600.00 \\ \hline
MNIST\_FC2 & 0.05 & 22 & -      & 600.00 & -      & 601.11 & -      & 600.11 & -      & 602.75 & -      & 600.00 \\ \hline
MNIST\_FC2 & 0.05 & 23 & -      & 600.00 & -      & 600.25 & -      & 600.10 & -      & 599.51 & -      & 600.00 \\ \hline
MNIST\_FC2 & 0.05 & 24 & -      & 600.00 & -      & 600.09 & -      & 600.10 & -      & 600.39 & -      & 600.00 \\ \hline
MNIST\_FC2 & 0.05 & 25 & -      & 600.00 & -      & 600.25 & -      & 600.10 & -      & 599.35 & -      & 600.00 \\ \hline
MNIST\_FC2 & 0.05 & 26 & sat   & 43.58  & sat   & 88.45  & -      & 600.10 & -      & 599.82 & sat   & 87.65  \\ \hline
MNIST\_FC2 & 0.05 & 27 & -      & 600.00 & -      & 600.14 & -      & 600.10 & -      & 601.04 & -      & 600.00 \\ \hline
MNIST\_FC2 & 0.05 & 28 & -      & 600.00 & -      & 600.35 & -      & 600.10 & -      & 599.27 & -      & 600.00 \\ \hline
MNIST\_FC2 & 0.05 & 29 & -      & 600.00 & -      & 600.01 & -      & 600.10 & -      & 600.22 & -      & 600.00 \\ \hline
MNIST\_FC2 & 0.05 & 30 & unsat & 8.33   & -      & 601.39 & -      & 600.11 & -      & 602.17 & -      & 600.00 \\ \hline
MNIST\_FC2 & 0.05 & 31 & -      & 600.00 & -      & 600.17 & unsat & 175.15 & unsat & 599.62 & -      & 600.00 \\ \hline
MNIST\_FC2 & 0.05 & 32 & unsat & 71.43  & -      & 600.49 & -      & 600.10 & -      & 602.71 & -      & 600.00 \\ \hline
MNIST\_FC2 & 0.05 & 33 & -      & 600.00 & -      & 600.29 & -      & 600.11 & -      & 599.53 & -      & 600.00 \\ \hline
MNIST\_FC2 & 0.05 & 34 & unsat & 284.36 & -      & 601.11 & -      & 600.11 & -      & 601.14 & -      & 600.00 \\ \hline
MNIST\_FC2 & 0.05 & 35 & -      & 600.00 & -      & 600.54 & -      & 600.11 & -      & 600.29 & -      & 600.00 \\ \hline
MNIST\_FC2 & 0.05 & 36 & -      & 600.00 & -      & 600.27 & -      & 600.10 & -      & 600.80 & -      & 600.00 \\ \hline
MNIST\_FC2 & 0.05 & 37 & -      & 600.00 & -      & 600.53 & unsat & 434.53 & unsat & 0.13   & -      & 600.00 \\ \hline
MNIST\_FC2 & 0.05 & 38 & unsat & 37.17  & -      & 601.19 & -      & 600.11 & -      & 603.79 & -      & 600.00 \\ \hline
MNIST\_FC2 & 0.05 & 39 & unsat & 74.23  & -      & 601.95 & -      & 600.11 & unsat & 0.20   & -      & 600.00 \\ \hline
MNIST\_FC2 & 0.05 & 40 & -      & 600.00 & -      & 600.11 & unsat & 409.12 & -      & 600.38 & -      & 600.00 \\ \hline
MNIST\_FC2 & 0.05 & 41 & unsat & 14.56  & -      & 601.60 & -      & 600.10 & -      & 601.87 & -      & 600.00 \\ \hline
MNIST\_FC2 & 0.05 & 42 & unsat & 317.77 & -      & 601.18 & -      & 600.10 & -      & 600.60 & -      & 600.00 \\ \hline
MNIST\_FC2 & 0.05 & 43 & -      & 600.00 & -      & 600.06 & unsat & 18.11  & -      & 600.04 & -      & 600.00 \\ \hline
MNIST\_FC2 & 0.05 & 44 & unsat & 9.55   & -      & 600.91 & -      & 600.10 & unsat & 0.13   & -      & 600.00 \\ \hline
MNIST\_FC2 & 0.05 & 45 & -      & 600.00 & -      & 600.22 & unsat & 90.01  & -      & 600.68 & -      & 600.00 \\ \hline
MNIST\_FC2 & 0.05 & 46 & -      & 600.00 & -      & 600.38 & -      & 600.11 & unsat & 0.23   & -      & 600.00 \\ \hline
MNIST\_FC2 & 0.05 & 47 & unsat & 6.82   & -      & 601.05 & -      & 600.10 & -      & 606.19 & -      & 600.00 \\ \hline
MNIST\_FC2 & 0.05 & 48 & -      & 600.00 & -      & 601.23 & -      & 600.10 & -      & 602.74 & -      & 600.00 \\ \hline
MNIST\_FC2 & 0.05 & 49 & sat   & 260.02 & -      & 600.01 & -      & 600.10 & -      & 599.78 & -      & 600.00 \\ \hline
MNIST\_FC2 & 0.05 & 50 & -      & 600.00 & -      & 600.08 & -      & 600.11 & -      & 600.41 & -      & 600.00 \\ \hline
MNIST\_FC3 & 0.05 & 1  & -      & 600.00 & -      & 600.92 & -      & 600.10 & -      & 664.75 & -      & 600.00 \\ \hline
MNIST\_FC3 & 0.05 & 2  & -      & 600.00 & -      & 600.36 & -      & 600.10 & -      & 599.77 & -      & 600.00 \\ \hline
MNIST\_FC3 & 0.05 & 3  & -      & 600.00 & -      & 601.64 & -      & 600.10 & -      & 600.26 & -      & 600.00 \\ \hline
MNIST\_FC3 & 0.05 & 4  & -      & 600.00 & -      & 602.70 & -      & 600.11 & -      & 600.24 & -      & 600.00 \\ \hline
MNIST\_FC3 & 0.05 & 5  & -      & 600.00 & -      & 600.72 & -      & 600.11 & -      & 602.35 & -      & 600.00 \\ \hline
MNIST\_FC3 & 0.05 & 6  & -      & 600.00 & -      & 601.17 & -      & 600.11 & -      & 601.89 & -      & 600.00 \\ \hline
MNIST\_FC3 & 0.05 & 7  & -      & 600.00 & -      & 605.97 & -      & 600.10 & -      & 604.94 & -      & 600.00 \\ \hline
MNIST\_FC3 & 0.05 & 8  & -      & 600.00 & -      & 605.35 & -      & 600.11 & -      & 602.60 & -      & 600.00 \\ \hline
MNIST\_FC3 & 0.05 & 9  & unsat & 127.18 & unsat & 0.28   & unsat & 159.94 & unsat & 0.15   & -      & 600.00 \\ \hline
MNIST\_FC3 & 0.05 & 10 & -      & 600.00 & -      & 601.57 & -      & 600.11 & -      & 600.05 & -      & 600.00 \\ \hline
MNIST\_FC3 & 0.05 & 11 & -      & 600.00 & -      & 600.67 & -      & 600.11 & -      & 598.97 & -      & 600.00 \\ \hline
MNIST\_FC3 & 0.05 & 12 & -      & 600.00 & -      & 604.37 & -      & 600.11 & -      & 600.07 & -      & 600.00 \\ \hline
MNIST\_FC3 & 0.05 & 13 & -      & 600.00 & -      & 603.92 & -      & 600.10 & -      & 600.02 & -      & 600.00 \\ \hline
MNIST\_FC3 & 0.05 & 14 & -      & 600.00 & -      & 601.61 & -      & 600.10 & -      & 618.54 & -      & 600.00 \\ \hline
MNIST\_FC3 & 0.05 & 15 & -      & 600.00 & -      & 604.49 & -      & 600.11 & -      & 601.17 & -      & 600.00 \\ \hline
MNIST\_FC3 & 0.05 & 16 & -      & 600.00 & -      & 601.61 & -      & 600.10 & -      & 599.65 & -      & 600.00 \\ \hline
MNIST\_FC3 & 0.05 & 17 & -      & 600.00 & -      & 602.09 & -      & 600.10 & -      & 601.13 & -      & 600.00 \\ \hline
MNIST\_FC3 & 0.05 & 18 & -      & 600.00 & -      & 600.99 & -      & 600.10 & -      & 600.77 & -      & 600.00 \\ \hline
MNIST\_FC3 & 0.05 & 19 & -      & 600.00 & -      & 600.99 & unsat & 187.84 & -      & 599.77 & -      & 600.00 \\ \hline
MNIST\_FC3 & 0.05 & 20 & -      & 600.00 & -      & 601.26 & -      & 600.10 & -      & 601.96 & -      & 600.00 \\ \hline
MNIST\_FC3 & 0.05 & 21 & -      & 600.00 & -      & 601.24 & -      & 600.11 & -      & 599.73 & -      & 600.00 \\ \hline
MNIST\_FC3 & 0.05 & 22 & -      & 600.00 & -      & 600.65 & -      & 600.10 & -      & 603.08 & -      & 600.00 \\ \hline
MNIST\_FC3 & 0.05 & 23 & -      & 600.00 & -      & 601.76 & -      & 600.11 & -      & 600.28 & -      & 600.00 \\ \hline
MNIST\_FC3 & 0.05 & 24 & -      & 600.00 & -      & 602.38 & -      & 600.10 & -      & 601.54 & -      & 600.00 \\ \hline
MNIST\_FC3 & 0.05 & 25 & -      & 600.00 & -      & 600.66 & -      & 600.10 & -      & 612.96 & -      & 600.00 \\ \hline
MNIST\_FC3 & 0.05 & 26 & sat   & 77.75  & sat   & 228.81 & -      & 600.11 & sat   & 219.04 & sat   & 600.00 \\ \hline
MNIST\_FC3 & 0.05 & 27 & -      & 600.00 & -      & 602.31 & -      & 600.11 & -      & 603.67 & -      & 600.00 \\ \hline
MNIST\_FC3 & 0.05 & 28 & -      & 600.00 & -      & 601.39 & -      & 600.10 & -      & 603.35 & -      & 600.00 \\ \hline
MNIST\_FC3 & 0.05 & 29 & -      & 600.00 & -      & 601.33 & -      & 600.11 & -      & 599.34 & -      & 603.12 \\ \hline
MNIST\_FC3 & 0.05 & 30 & -      & 600.00 & -      & 603.88 & -      & 600.11 & -      & 608.68 & -      & 600.00 \\ \hline
MNIST\_FC3 & 0.05 & 31 & -      & 600.00 & -      & 601.24 & -      & 600.10 & -      & 602.72 & -      & 600.00 \\ \hline
MNIST\_FC3 & 0.05 & 32 & -      & 600.00 & -      & 600.81 & -      & 600.11 & -      & 601.27 & -      & 600.00 \\ \hline
MNIST\_FC3 & 0.05 & 33 & -      & 600.00 & -      & 601.65 & unsat & 238.14 & -      & 599.81 & -      & 600.00 \\ \hline
MNIST\_FC3 & 0.05 & 34 & unsat & 177.44 & unsat & 0.29   & -      & 600.10 & unsat & 0.15   & -      & 600.00 \\ \hline
MNIST\_FC3 & 0.05 & 35 & -      & 600.00 & -      & 601.70 & -      & 600.10 & -      & 601.81 & -      & 600.00 \\ \hline
MNIST\_FC3 & 0.05 & 36 & -      & 600.00 & -      & 600.29 & -      & 600.10 & -      & 606.02 & -      & 600.00 \\ \hline
MNIST\_FC3 & 0.05 & 37 & unsat & 557.04 & unsat & 34.23  & -      & 600.11 & unsat & 0.15   & -      & 600.00 \\ \hline
MNIST\_FC3 & 0.05 & 38 & -      & 600.00 & -      & 601.20 & -      & 600.10 & -      & 604.42 & -      & 600.00 \\ \hline
MNIST\_FC3 & 0.05 & 39 & -      & 600.00 & -      & 603.12 & -      & 600.10 & -      & 599.62 & -      & 600.00 \\ \hline
MNIST\_FC3 & 0.05 & 40 & -      & 600.00 & -      & 602.38 & -      & 600.10 & -      & 599.31 & -      & 600.00 \\ \hline
MNIST\_FC3 & 0.05 & 41 & -      & 600.00 & -      & 600.08 & -      & 600.10 & -      & 612.55 & -      & 618.24 \\ \hline
MNIST\_FC3 & 0.05 & 42 & unsat & 193.70 & -      & 604.19 & -      & 600.10 & -      & 606.26 & -      & 618.24 \\ \hline
MNIST\_FC3 & 0.05 & 43 & -      & 600.00 & -      & 601.62 & -      & 600.10 & -      & 602.42 & -      & 600.00 \\ \hline
MNIST\_FC3 & 0.05 & 44 & -      & 600.00 & -      & 600.71 & -      & 600.11 & -      & 601.50 & -      & 600.17 \\ \hline
MNIST\_FC3 & 0.05 & 45 & -      & 600.00 & -      & 603.13 & -      & 600.10 & -      & 602.57 & -      & 620.11 \\ \hline
MNIST\_FC3 & 0.05 & 46 & -      & 600.00 & -      & 600.64 & -      & 600.10 & -      & 602.45 & -      & 616.95 \\ \hline
MNIST\_FC3 & 0.05 & 47 & unsat & 165.03 & unsat & 20.46  & -      & 600.11 & unsat & 0.34   & -      & 600.17 \\ \hline
MNIST\_FC3 & 0.05 & 48 & -      & 600.00 & -      & 601.24 & -      & 600.10 & -      & 599.04 & -      & 616.95 \\ \hline
MNIST\_FC3 & 0.05 & 49 & -      & 600.00 & -      & 600.28 & -      & 600.10 & -      & 600.30 & -      & 600.17 \\ \hline
MNIST\_FC3 & 0.05 & 50 & -      & 600.00 & -      & 601.39 & -      & 600.10 & -      & 599.45 & -      & 600.17 \\ \hline
\end{longtable}
\end{document}